\documentclass[letterpaper]{article} 
\usepackage{aaai25}  
\usepackage{times}  
\usepackage{helvet}  
\usepackage{courier}  
\usepackage[hyphens]{url}  
\usepackage{graphicx} 
\usepackage{natbib}  
\usepackage{caption} 
\usepackage{algorithm}
\usepackage{algorithmic}
\usepackage{amsmath}

\usepackage{newfloat}
\usepackage{listings}
\DeclareCaptionStyle{ruled}{labelfont=normalfont,labelsep=colon,strut=off} 
\floatstyle{ruled}
\newfloat{listing}{tb}{lst}{}
\floatname{listing}{Listing}

\usepackage{makecell}
\usepackage{booktabs}
\usepackage{multirow}
\usepackage{xspace}
\newcommand{\ours}{K-ON\xspace}

\usepackage{amsmath,amsfonts,bm}

\def\eqref#1{equation~\ref{#1}}

\def\1{\bm{1}}

\def\rvh{{\mathbf{h}}}

\def\rvp{{\mathbf{p}}}

\def\rmA{{\mathbf{A}}}
\def\rmB{{\mathbf{B}}}

\def\rmL{{\mathbf{L}}}
\def\rmM{{\mathbf{M}}}

\def\rmP{{\mathbf{P}}}

\def\rmW{{\mathbf{W}}}

\DeclareMathAlphabet{\mathsfit}{\encodingdefault}{\sfdefault}{m}{sl}
\SetMathAlphabet{\mathsfit}{bold}{\encodingdefault}{\sfdefault}{bx}{n}

\def\gE{{\mathcal{E}}}

\def\gG{{\mathcal{G}}}
\def\gH{{\mathcal{H}}}

\def\gM{{\mathcal{M}}}
\def\gN{{\mathcal{N}}}

\def\gP{{\mathcal{P}}}

\def\gR{{\mathcal{R}}}

\def\gT{{\mathcal{T}}}

\def\gV{{\mathcal{V}}}

\newcommand{\Ls}{\mathcal{L}}
\newcommand{\R}{\mathbb{R}}

\newcommand{\KL}{D_{\mathrm{KL}}}

\title{K-ON: Stacking Knowledge On the Head Layer of Large Language Model}
\author {
    Lingbing Guo\textsuperscript{\rm 1,2},
    Yichi Zhang\textsuperscript{\rm 1,2},
    Zhongpu Bo\textsuperscript{\rm 3},
    Zhuo Chen\textsuperscript{\rm 1,2},
    Mengshu Sun\textsuperscript{\rm 3},\\
    Zhiqiang Zhang\textsuperscript{\rm 3},
    Wen Zhang\textsuperscript{\rm 4,2}\thanks{Corresponding Authors.},
    Huajun Chen\textsuperscript{\rm 1,2,5*}
}
\affiliations {
    \textsuperscript{\rm 1}College of Computer Science and Technology, Zhejiang University\\
    \textsuperscript{\rm 2}ZJU-Ant Group Joint Lab of Knowledge Graph\\
    \textsuperscript{\rm 3}Ant Group\\
    \textsuperscript{\rm 4}School of Software Technology, Zhejiang University\\
    \textsuperscript{\rm 5}Zhejiang Key Laboratory of Big Data Intelligent Computing\\
    \{lbguo, zhangyichi2022, zhang.wen, huajunsir\}@zju.edu.cn
}

\begin{document}

\maketitle

\begin{abstract}
Recent advancements in large language models (LLMs) have significantly improved various natural language processing (NLP) tasks. Typically, LLMs are trained to predict the next token, aligning well with many NLP tasks. However, in knowledge graph (KG) scenarios, entities are the fundamental units and identifying an entity requires at least several tokens. This leads to a granularity mismatch between KGs and natural languages. To address this issue, we propose K-ON, which integrates KG knowledge into the LLM by employing multiple head layers for next $k$-step prediction. K-ON can not only generate entity-level results in one step, but also enables contrastive loss against entities, which is the most powerful tool in KG representation learning. Experimental results show that K-ON outperforms state-of-the-art methods that incorporate text and even the other modalities.
\end{abstract}

\section{Introduction}
\label{sec:intro}
Large language models (LLMs) are trained on vast amounts of corpora and store world knowledge within billions of neurons~\cite{chatgpt,llama-2}. Despite of the prosperity in LLM-based applications, unleashing its power for knowledge graph (KG) tasks remains challenging. 

Tokens are the basic elements for language models, but it needs to take at least several tokens to describe and identify different entities in a KG. Creating new token identifiers for each entity and learning them during fine-tuning is an alternative choice, however, which is extremely time-consuming and may negatively affect the native performance of LLMs.

\begin{figure}[t]
    \centering
    \includegraphics[width=\linewidth]{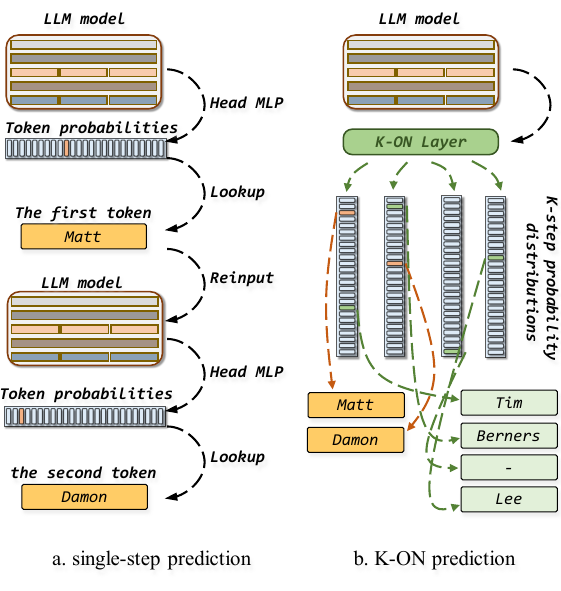}
    \caption{A comparison of single-step prediction and the proposed \ours prediction. Left: in the conventional single-step prediction, obtaining an output of an entity necessitates recurrently feeding input data and cannot be parallelized across different entities. Right: the \ours prediction generates an entity in a single step and allows for parallelization across multiple entities, thereby enabling entity-level contrastive learning. }
    \label{fig:example}
\end{figure}

In this paper, we explore how to effectively and efficiently use multiple tokens to describe entities in a given KG. Evidently, directly optimizing the sequence prediction objective will results in the out-of-KG problem as LLM lacks awareness of the KG’s entities, while enumerating all entities in the input instruction is unrealistic. Take Figure~\ref{fig:example}a as an example, the task is to predict the target entity \textit{Matt Damon} given the incomplete triplet (\textit{The Bourne Identity (2002 film), starring, ?}). The vanilla learning schema is inefficient because generating a single entity requires multiple steps and cannot be parallelized across entities.

Most existing methods compromise on this dilemma and apply LLMs only to simple KG tasks ~\cite{kgllm1,kgllm2,kgllm4,kgllm3}, such as verifying the correctness of a triplet ~\cite{KoPa} or predicting the target from a limited number of candidates ~\cite{KGllama}. In contrast, we propose \emph{\ours} to employ $K$ head layers for predicting entities at one shot and stack knowledge on these heads by entity-level contrastive learning.

As shown in Figure~\ref{fig:example}b, \ours adapts $K$ different head layers from the original LLM head, where the $k$-th head is responsible for predicting the $k$-th token for all entities. For example, an entity \textit{Matt Damon} is tokenized into $K$ input token IDs $t_{0:K-1}$, with $t_0$ representing \textit{Matt} and $t_{K-1}$ being the last token \textit{Damon} or padding token. We then extract the the probability of the first token from the first head layer, and so forth. For the other entities that may serve as negative examples in contrastive learning, we can reuse the $K$-step probability estimations to extract their scores.

The risk underlying the prediction of the next $K$ tokens is \emph{over-optimization}. The model may over-optimize for predicting just the next $K$ tokens, dropping the fact that these tokens form the target entity. For instance, while minimizing cross-entropy loss for the first token, with \textit{Matt} as the positive label, many negative entries (tokens) are not constituent elements of an entity. Moreover, increasing the probability of \textit{Matt} does not always equate to maximizing the probability of \textit{Matt Damon}. To tackle this issue, \ours employs an entity-level contrastive loss, which treats the $K$-step predictions as an integrity and estimates the joint probabilities.

Another risk in next-$K$-token prediction is \emph{distribution corruption}. In the original schema, the probability of the second token ``\textit{Damon}'' is conditioned on the first token ``\textit{Matt}''. However, this conditioning is absent in the next-$K$-tokens schema. Such discrepancy may degrade the performance as well as inference ability of the original LLM. 

To address this issue, we propose \emph{head trajectory tuning (HTT)} to align the distribution trajectories between the original LLM's prediction and the next-$K$-token predictions. We first leverage a conditional attention layer to process the hidden states from $K$ head layers to reconstruct the sequential dependencies between different steps. Then, we compute a standard sequence prediction loss with the original LLM head layer as target. Finally, we can align the output probability sequence of our next-$k$-token predictions with the original estimates by minimizing their KL divergence. In this way, we expect to mimic the single-step prediction process using a set of learnable functions within \ours.

We evaluate \ours on the KG completion task without any simplification on the task setting. Our experiments demonstrate that \ours not only outperforms the conventional methods, but also achieves better performance than the multi-modal methods that leverage additional textual and visual information. Furthermore, \ours is also an efficient method: although incorporating the LLM requires more GPU resources, the number of training epochs is reduced from 1,000 to 5 compared with conventional methods. The overall fine-tuning time is less than $1$ hour on the DB15K dataset with $8$ A100 GPUs.

\begin{figure*}[thb]
    \centering
    \includegraphics[width=\textwidth]{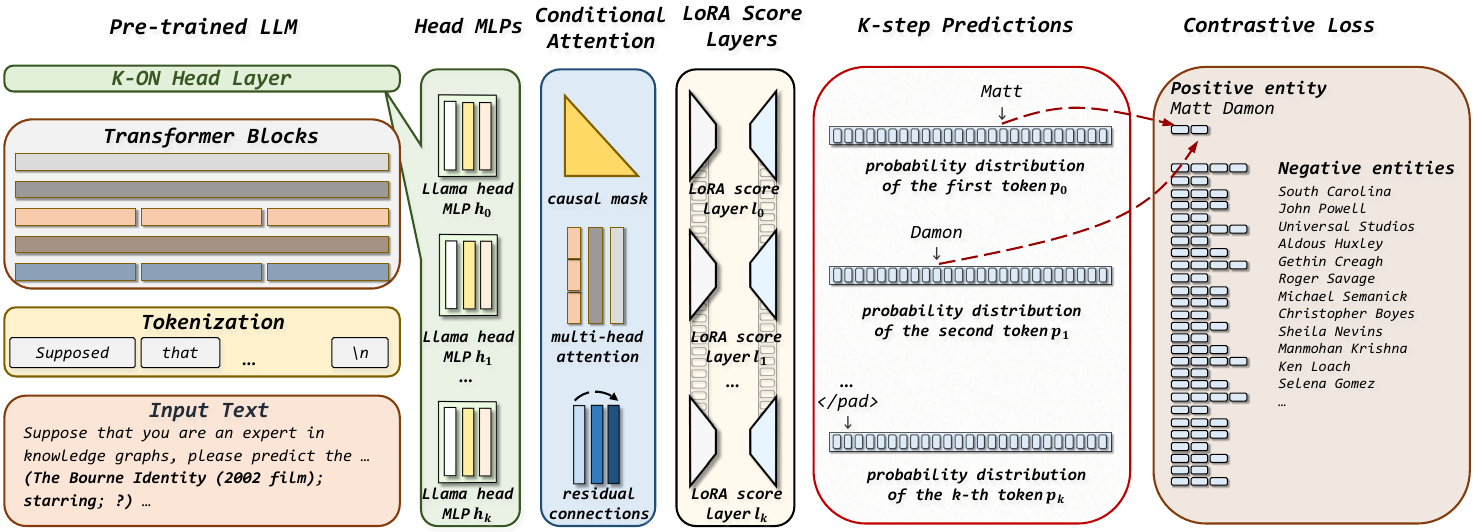}
    \caption{Overview of the \ours architecture. From left to right: (1) The LLM processes the input text containing incomplete triplet information; (2) The resulting hidden states are input to distinct head MLPs within \ours; (3) A compact conditional Transformer refines the corresponding outputs to capture sequential dependencies; (4) LoRA score layers are employed to transform the hidden states into $K$ probability distribution estimations; (5-6) Aggregating the elements from the respective probability vectors, \ours computes the probabilities of all candidate entities simultaneously.}  
    \label{fig:arch}
\end{figure*}

\section{Related Works}
\label{sec:related}

\paragraph{Knowledge Graph Completion}
Knowledge Graph (KG) completion is one of the most important tasks in the KG area. Conventional methods leverage triplet information as training data but often ignore the rich contextual information embedded within the text~\cite{TransE,ConvE,RSN,CompGCN,decentRL,HittER,DET,NeoEA}. In other words, they assume that the entities carry no self-feature (including their names, e.g., \textit{Matt Damon}), and the relational connections are the only informative source. Recently, methods leveraging text and image information have proposed and achieved state-of-the-art performance on many benchmarks~\cite{xie_image-embodied_2017-IKRL,wang_multimodal_2019-TransAE,KG-Bert,KGLM,FLT-LM,DBLP:conf/ijcnn/ZhangCZ23-MANS,GEEA}. They can be divided into two groups: one focuses on the integration of more modalities~\cite{DBLP:journals/apin/LuWJHL22-MMKRL,lee_vista_2023-VISTA,MAT}; and the other concentrates on the fine-tuning of language models to better encoding text information~\cite{KG-Bert,KGLM,FLT-LM}. 

\paragraph{LLM-based Knowledge Graph Completion}
Entities possess rich features often represented in text forms such as descriptions, tables, and attributes. Many (mostly multi-modal) methods propose leveraging language models to encode text information and use the resulting representations for prediction~\cite{KG-Bert,StAR,FLT-LM,KGLM,Meaformer,DBLP:conf/sigir/ZhangCGXHLZC24,DBLP:journals/corr/abs-2405-16869,DBLP:journals/corr/abs-2410-07526}. In the recent advances, most LLM-based methods can be classified into this category. For example, LLMKGC \cite{llmkgc-ningyu} directly feeds the textual triplets to ChatGPT~\cite{chatgpt} for KG completion, although the results are not very promising. KGLlama~\cite{KGllama} and KoPA~\cite{KoPa} fine-tune LLMs on triplet verification, i.e., estimating the correctness of a given triplet. These LLM-based methods employ in-context learning or LoRA-based fine-tuning~\cite{icl1,icl2}. To our knowledge, no prior work has explored integrating KGs into the head layer of LLMs. The current LLM-based methods leverage additional text information but are directly compared against conventional methods~\cite{KGllama,KoPa,kicgpt}, which may lead to unfair evaluations. Therefore, in this paper, we consider multi-modal datasets as benchmarks~\cite{MMKG,MMRNS,DBLP:journals/corr/abs-2402-05391}.

\paragraph{Multi-Head LLMs}
There are several works employing multiple head layers in LLMs. Medusa~\cite{medusa} proposes a tree attention mechanism for multi-step training and inference. The $K$ different head layers are initialized with the original weights and then fine-tuned independently. MultiToken~\cite{multi-head} discovers that training LLMs and multiple head layers from scratch can outperform the single-head version, with this advantage being more significant in larger models. Unlike these methods, the $K$ head layers in our approach are not only used for generating multiple future tokens in one step but also confine the output space to KGs and enable entity-level contrastive learning. Our work explores a new direction for integrating LLMs with KGs.

\section{Methodology}
\label{sec:method}
In this section, we present the details of \ours. We begin with a preliminary overview of knowledge graphs and large language models, and then illustrate the architecture and implementation of \ours. Finally, we introduce head trajectory tuning as a self-supervised optimization method for \ours. 

\subsection{Preliminaries}

\paragraph{Knowledge Graphs} We describe a knowledge graph by $\gG=\{\gE, \gR, \gT\}$, where $\gE$, $\gR$, $\gT$ denote the entity, relation, and triplet sets, respectively. As one of the most important tasks in KG area, KG completion aims to predict the missing entity given an incomplete triplet~\cite{TransE}, i.e., predicting a tail entity $e_2$ given $(e_1, r_1, ?)$ or predicting the head entity $e_1$ given $(?, r_1, e_2)$.

\paragraph{Large Language Models} Generally, a large language model comprises the following components: a tokenizer, which splits the input text into a sequence of $N$ tokens $t_{0:N-1} =  \{t_n | t_n \in \gV \}_{n=0}^{N-1}$ with $|\gV|\geq N$ being the vocabulary; a Transformer-based model $\gM$, which processes the token sequence and generates a corresponding sequence of hidden states for prediction;
\begin{align}
\label{eq:llm-hidden}
    \rvh^m_{0:N-1} = \gM(t_{0:N-1});
\end{align}
and a head layer $\gH$, which maps each hidden state to a probability distribution $\rvp_n \in \R^{|\gV|}$:
\begin{align}
    \rvp_n = \gH(\rvh^m_n).
\end{align}
In this paper, we focus primarily on the head layer of the LLM, and demonstrate that integrating knowledge graphs into the LLM at only this stage is sufficient to achieve state-of-the-art performance in KG completion.

\subsection{\ours} 
Entities typically require multiple tokens to be identifiable by name, which introduces a discrepancy between the entity distribution and the token probability distribution during prediction. Specifically, in a standard fine-tuning schema, the prediction objective is optimized at the token level rather than at the entity level. Consequently, the LLM is unaware of which entities are present in the given KG, except those provided in the context.

Given the vast number of possible entities, it is impractical to include all candidate entities in the input text. An alternative approach worth exploring is manipulating the output probability distributions. If we can obtain the probability sequence of the constituent tokens for each entity, we can construct an entity-level entropy-based loss for the LLM. For example, suppose that we have constructed the input query containing the information of $e_1, r_1, ?$, we wish that the LLM precisely generates the output, $e_2$, which comprises maximally $K$ tokens. Then, we can extract the probability of each token and combine them as the joint probability for $e_2$. However, achieving this is challenging, primarily due to computational costs. Iteratively feeding every negative example into the LLM for back-propagation is computationally intensive. As shown in Figure~\ref{fig:example}a, it is not parallelizable across entities.

To address this problem, we propose \ours. Figure~\ref{fig:arch} illustrates the overall architecture of \ours. In addition to the original LLM's input and Transformer layers, \ours introduces five new modules to support $K$-step token prediction.

\paragraph{Head MLPs}
We first employ multiple head MLPs to process the output hidden states of the LLM into inputs for different steps. Specifically, each MLP consists of three components: a fully-connected layer $\rmW^h_k \in \R^{d\times d}$, an activation function $\sigma$, and a normalization layer $\rmL^h_k$:
\begin{align}
    \label{eq:head_mlp}
    \rvh^h_{0:K-1} = \{\rmL^h_k( \sigma(\rmW^h_k \rvh^m_0) ) \}_{k=0}^{K-1},
\end{align}
where $\rvh^h_{0:K-1}$ are the $K$ output hidden states of the head MLPs. It is worth noting that the LLM uses a decoder-only architecture, and the input hidden state $\rvh_0^m$ for \ours is the last output hidden state of the query text. The task is to follow the query text and generate $K$ subsequent tokens as predictions of entities. Similar to Llama 2~\cite{llama-2}, we use SiLU~\cite{SiLU} and LlamaRMSNorm~\cite{RMSNorm} as the activation function $\sigma$ and normalization layer $\rmL^h_k$, respectively. The bias vector is not used.

\paragraph{Conditional Attention}
While the head MLPs in \ours are independent of each other, the subsequent outputs in the original LLM are conditioned on the previous inputs. Therefore, we leverage a small Transformer $\gM_s$ to mimic this process by incorporating a causal mask $\rmM \in \R^{K\times K}$:
\begin{align}
    \rmM_{ij} = 
    \begin{cases}
        1  \quad i\geq j,\\
        0 \quad i < j,
    \end{cases}
\end{align}
where $\rmM_{ij}$ indicates the value at the $i$-th row and $j$-th column of $\rmM$. When processing the $k$-th step, only the outputs of the previous $k-1$ steps from the head MLPs can be observed for the Transformer.

Another noteworthy aspect of the conditional attention is the residual connection layer. Specifically, we add the attention output to the initial output of the LLM to produce the final output:
\begin{align}
    \rvh^a_k = \gM_s(\rvh^h_{0:k}, \rmM) + \rvh^m_0,
\end{align}
where $\gM_s$, $\rmM$ are aforementioned small Transformer and causual mask, respectively. With the residual connection, we can initialize the head MLPs with zeros, allowing them to gradually learn the adaptation for $K$-step prediction from the initial LLM hidden $\rvh^m_0$.

\paragraph{LoRA Score Layer}
We use different score layers to estimate the probability distribution for each step. Unlike \cite{multi-head} which trains each score layer from scratch, we propose to use a low-rank adaptation (LoRA)~\cite{lora} layer for each step. This can be expressed as:
\begin{align}
    \rmW_k^S &= \rmW^S + \rmA_k \rmB_k\\
    \rvp_k &= \rmW_k^S \rvh^a_k
\end{align}
where $\rmW^S \in \R^{|\gV| \times d}$ is the original score layer of the LLM. $|\gV|$ denotes the vocabulary size (number of tokens) of the LLM. $\rmA_k \in \R^{d\times r}$ and $B_k \in \R^{r\times d}$ are the down-scaling and upper-scaling matrices in LoRA. The hyper-parameter $r \ll d$ is the reduced dimensionality. Before fine-tuning, $\rmA_k$ is initialized randomly while $\rmB_k$ is initialized to zero. This ensures that the output of the adaptation layers is identical to the original head layer.

\paragraph{$K$-step Gathering}
After obtaining the $K$-step predictions $\rvp_0, \rvp_1, ..., \rvp_{K-1}$, we need to efficiently extract the elements relevant to each entity. To achieve this, we first convert each entity to an identifiable sequence of tokens:
\begin{align}
\label{eq:k-step-label}
    t_{0:K-1}^e = \gP(\tau(l^e), K)
\end{align}
where $l^e$ is the textual label of the entity $e$ and $\tau$ is the tokenizer. We perform padding and truncation $\gP$ to ensure that all entity token sequences have the same length $K$.

Next, we stack the $K$-step predictions $\rvp_0, \rvp_1, ..., \rvp_{K-1}$ into a probability matrix $\rmP$:
\begin{align}
    \rmP = 
    \begin{pmatrix}
        \rvp_0 & \rvp_1 & \dots & \rvp_{K-1}
    \end{pmatrix}^T,
\end{align}
such that the each token probability $p_k^e$ can be extracted from $\rmP$ with $t_k^e$ as indices:
\begin{align}
\label{eq:k-on-prob}
    \rvp^e = \rmP_{0, t_0^e}, \rmP_{1, t_1^e}, ..., \rmP_{K-1, t_{K-1}^e}.
\end{align}
Here, $\rvp^e$ is the token probability sequence of entity $e$, with a strict length of $K$.

\begin{algorithm}[t]
	\caption{\ours for KG Completion}
	\label{alg:k-on}
	\begin{algorithmic}[1]
		\STATE {\bfseries Input:} the training KG $\gG$, the language model $\gM$, the \ours head layers $\gH_{0:K-1}$;
		\FOR{{\bfseries each} batched triplets in the training KG $\gG$} 
        \STATE Construct and tokenize the input queries; create the $K$-step labels for target and negative entities (Equation~(\ref{eq:k-step-label}));
        \STATE $\rvh^m_0 \leftarrow \gM(t_{0:N-1})$, obtaining the output hidden states of LLM (Equation~(\ref{eq:llm-hidden}));  
        \STATE $\rvp^{e} \leftarrow \gH_{0:K-1}(\rvh^m_0)$, estimating \ours predictions following Equations (\ref{eq:head_mlp}-\ref{eq:k-on-prob});
        \STATE Compute entity-level contrastive loss $\Ls_\text{NCE}(e)$ (Equation~(\ref{eq:entity-contrastive-loss}));
        \STATE Compute the supervised fine-tuning loss  $\Ls_\text{sft}(e)$ (Equation~(\ref{eq:sft}));
        \STATE Compute the token distribution tuning loss  $\Ls_\text{tdt}(e)$ (Equation~(\ref{eq:tdt}));
		\STATE Jointly minimizing all three losses;
        \ENDFOR
	\end{algorithmic}
\end{algorithm}

\paragraph{Contrastive Loss}
To incorporate the entity-level contrastive loss, we first estimate the scalar probability $p^e$ from its token probability sequence $\rvp^e$. Our experimental results indicates that a weighted sum achieves the best performance:
\begin{align}
\label{eq:entity-contrastive-loss}
    p^e = \sum_{p_k \in \rvp^e} \alpha_k p_k,
\end{align}
where $\alpha_k$ is a learnable weight for $k$-th step shared across different entities.

By using the above equation to gather the scalar estimates for both positive and negative entities, we can construct an effective contrastive loss. Specifically, we randomly sample entities from the entity set $\gE$ to form negative examples $\gN = \{e_j|e_j \neq e, e_j \in \gE \}$:
\begin{align}
    \Ls_\text{NCE}(e) = -\log p^e + \frac{1}{|\gN|} \sum_{e_j \in \gN} \log p^{e_j},
\end{align}
where $p^e$ and $p^{e_j}$ denote the joint probabilities for the positive entity $e$ and negative entity $e_j$, respectively.
\begin{table}[t]
\centering
\resizebox{\columnwidth}{!}{
\begin{tabular}{l|rrrrrrr}
\toprule
Dataset   & \# Entity    & \# Relation   & \#Train & \#Valid & \#Test & \# Text & \# Image \\ \midrule
DB15K & 12,842 & 279 & 79,222  & 9,902   & 9,904  & 12,842 & 12,818  \\
MKGW    & 15,000 & 169  & 34,196   & 4,276    & 4,274  & 14,123 & 14,463 \\
\bottomrule
\end{tabular}
}
\caption{Statistics of the datasets.}
\label{tab:dataset}
\end{table}

\begin{table*}[!t]
	\centering
	\resizebox{\textwidth}{!}{\scriptsize
	\begin{tabular}{lccccccccc}
		\toprule
		\multirow{2}{*}{Model} & \multirow{2}{*}{Modality}  &  \multicolumn{4}{c}{DB15K} & \multicolumn{4}{c}{MKGW} \\ \cmidrule(lr){3-6} \cmidrule(lr){7-10}
		& & MRR$\uparrow$ & Hits@1$\uparrow$ & Hits@3$\uparrow$ & Hits@10$\uparrow$ & MRR$\uparrow$ & Hits@1$\uparrow$  & Hits@3$\uparrow$ & Hits@10$\uparrow$\\ \midrule
		TransE~\cite{TransE}  & S & 24.86 & 12.78 & 31.48 & 47.07 & 29.19 & 21.06 & 33.20 & 44.23 \\
        DistMult~\cite{DistMult} & S & 23.03 & 14.78 & 26.28 & 39.59 & 20.99 & 15.93 & 22.28 & 30.86 \\
		RotatE~\cite{RotatE} & S   & 29.28 & 17.87 & 36.12 & 49.66 & 33.67 & 26.80 & 36.68 & 46.73 \\
        \midrule
        IKRL \cite{xie_image-embodied_2017-IKRL} & S+I & 26.82 & 14.09 & 34.93 & 49.09 & 32.36 & 26.11 & 34.75 & 44.07\\
        TransAE \cite{wang_multimodal_2019-TransAE} & S+I & 28.09 & 21.25 & 31.17 & 41.17 & 30.00 & 21.23 & 34.91 & 44.72\\
        KG-Bert~\cite{KG-Bert} & S+T & 23.94 & 11.98 & 31.05 & 46.54 & 28.68 & 21.12 & 32.57 & 43.46 \\
        MMKRL \cite{DBLP:journals/apin/LuWJHL22-MMKRL} & S+T+I & 26.81 & 13.85 & 35.07 & 49.39 & 30.10 & 22.16 & 34.09 & 44.69\\
        OTKGE \cite{cao_otkge_2022-OTKGE} & S+T+I & 23.86 & 18.45 & 25.89 & 34.23 & 34.36 & \underline{28.85} & 36.25 & 44.88\\
        MMRNS \cite{MMRNS} & S+T+I & 32.68 & 23.01 & 37.86 & 51.01 & \underline{35.03} & 28.59 & 37.49 & \underline{47.47}\\
        KGLM~\cite{KGLM} & S+T & 28.47 & 17.66 & 36.02 & 48.89 & 34.12 & 27.01 & 36.87 & .46.62\\
        QEB \cite{DBLP:conf/mm/WangMCML023-TIVA} & S+T+I & 28.18 & 14.82 & 36.67 & 51.55 & 32.38 & 25.47 & 35.06 & 45.32\\
        VISTA \cite{lee_vista_2023-VISTA} & S+T+I & 30.42 & 22.49 & 33.56 & 45.94 & 32.91 & 26.12 & 35.38 & 45.61\\
        MANS \cite{DBLP:conf/ijcnn/ZhangCZ23-MANS} & S+T+I & 28.82 & 16.87 & 36.58 & 49.26 & 30.88 & 24.89 & 33.63 & 41.78\\
        FLT-LM~\cite{FLT-LM} & S+T & 33.45 & 24.56 & 37.67 &  50.12 & 32.75 & 25.89 & 32.87 & 44.56\\
        AdaMF \cite{MAT} & S+T+I & 32.51 & 21.31 & \underline{39.67} & \underline{51.68} & 34.27 & 27.21 & \underline{37.86} & 47.21\\
        \midrule
        KG-Llama-7b~\cite{KGllama} & S+T & - & 13.46 & - & - & - & 20.20 & - & - \\
        GPT 3.5 Turbo~\cite{llmkgc-ningyu} & S+T & - & 21.71 & - & - & - & 22.66 & - & - \\
        \midrule
        \ours & S+T & \textbf{38.10} & \textbf{30.13} & \textbf{42.77} & \textbf{53.59} & \textbf{36.64} & \textbf{30.05} & \textbf{38.72} & \textbf{48.26}\\
		\bottomrule
	\end{tabular}}
        \caption{The main KG completion results. The best and second-best results are \textbf{boldfaced} and \underline{underlined}, respectively. S, T, I indicate structure, text, and image, respectively. $\uparrow$: higher is better; $\downarrow$: lower is better. -: unavailable entry. }
	\label{tab:main}
\end{table*}

\subsection{Head Trajectory Tuning}
Entity-level contrastive learning is optimized against entities, which may inadvertently disrupt the token-level predictions of the original LLM, affecting performance on both common and training corpora. To mitigate this issue, we propose head trajectory tuning (HTT) to align the sequence estimations between single-step and $K$-step predictions.

\paragraph{Supervised Fine-Tuning}
HTT consists of two objectives. The first is tuning the LLM on the training corpus, also known as supervised fine-tuning (SFT). For this, we apply LoRA to the LLM model and optimize the single-step estimations against the ground truth:
\begin{align}
\label{eq:sft}
    \Ls_\text{sft}(e) = \sum_{k=0}^{K-1} \Big(-\log p_k^e + \frac{1}{\gV} \sum_{e_j \in \gV} \log p^{e_j}_k\Big),
\end{align}
where $\Ls_\text{sft}(e)$ is the supervised fine-tuning loss, with $p_k^e$ denoting the target probability at the $k$-th step and $\gV$ representing the token vocabulary.

\paragraph{Token Distribution Tuning}
Then, we propose token distribution tuning to align the probability estimations of \ours with those of the original LLM. Specifically, we minimize the KL-divergence~\cite{KLD} between each pair of estimations along the output trajectories.
\begin{align}
\label{eq:tdt}
    \Ls_\text{tdt}(e) = \sum_{k=0}^{K-1} \KL(\rvp_k^\text{e, k-on}, \rvp_k^\text{e, llm}),
\end{align}
where $p_k^\text{e, k-on}$, $p_k^\text{e, llm}$ denote the $k$-th token probability distributions of \ours and the original LLM head, respectively.

\subsection{Implementation}
We present Algorithm~\ref{alg:k-on} to illustrate the implementation of \ours step by step. We first construct the input text for each triplet in the training set, and then use the tokenizer to convert the query into token IDs. We feed the input into the LLM to obtain the hidden states, which will be estimated by \ours and the original head layer, respectively. Lastly, we compute the losses introduced before, and jointly minimize them until convergence.

\section{Experiment}
\label{sec:expr}
In this section, we conduct experiments to verify the effectiveness of the proposed \ours. 

\subsection{Setting}
We employ Llama-2-chat-7B~\cite{llama-2} as the base LLM model and train \ours with $8$ A100 GPUs. The learning rate is set to $1\mathrm{e}{-4}$ in all experiments, and we use AdamW~\cite{Adam} as the optimizer. The batch-size per device is set to $12$ and the gradient accumulation is set to $8$ to obtaining a larger batch-size. We follow~\cite{TransE,DBLP:journals/apin/LuWJHL22-MMKRL,MAT} to report the MRR and Hits@K results with filtered ranks.

We consider various KG completion methods as baselines: the conventional structure-only methods, such as TransE~\cite{TransE} and RotatE~\cite{RotatE}; the methods leveraging image information, such as IKRL~\cite{xie_image-embodied_2017-IKRL} and TransAE~\cite{wang_multimodal_2019-TransAE}; the methods leveraging text information, such as KG-Bert~\cite{KG-Bert} and FLT-LM~\cite{FLT-LM}; the methods leveraging both text and image information, such as MMKRL~\cite{DBLP:journals/apin/LuWJHL22-MMKRL} and MANS~\cite{DBLP:conf/ijcnn/ZhangCZ23-MANS}; and the LLM-based methods KG-Llama-7b~\cite{KGllama} and GPT 3.5~\cite{llmkgc-ningyu};

\subsection{Datasets}
We consider DB15K and MKGW as benchmark, which are widely used in many recent works~\cite{xie_image-embodied_2017-IKRL,MMRNS,lee_vista_2023-VISTA,DBLP:conf/ijcnn/ZhangCZ23-MANS,MAT}. The two datasets include not only the structural triplet data, but also the rich information of text and others. Thereby, we believe conducting experiments on them can gain a more comprehensive understanding on different methods and ensure a fairer comparison. The statistics of these two datasets are shown in Table~\ref{tab:dataset}.

\subsection{Results}
The main experimental results are shown in Table~\ref{tab:main}. We can find that the methods considering additional information generally perform better than the structure-only methods, which verifies the effectiveness of leveraging external informative sources. Among these methods, the proposed \ours achieves the best results, significantly surpassing all baseline methods.

However, we also observe that the performance gap between conventional methods and multi-modal approaches narrows on the MKGW dataset. For instance, FLT-LM~\cite{FLT-LM} and MANS~\cite{DBLP:conf/ijcnn/ZhangCZ23-MANS} perform worse than RotatE on the MKGW dataset. This suggests that while additional information can be beneficial, it may also introduce noise, which does not always aid in entity prediction. Thanks to advancements in language models, \ours leverages LLM to process text information and is optimized against an entity-level contrastive loss. Consequently, it still significantly outperforms all baselines across all metrics on the MKGW dataset.

The existing LLM-based methods, as mentioned in previous sections, are optimized against tokens rather than entities. Thus, their performance on the standard KG completion task is generally unsatisfactory, falling significantly short of many non-LLM methods. This phenomenon has also been examined by ~\cite{llmkgc-ningyu}. In contrast, our \ours introduces entity-level contrastive loss, enabling the LLM to more effectively explore the KG structure.

\begin{table}[t]
	\centering
	\resizebox{\linewidth}{!}{\scriptsize
	\begin{tabular}{l|cccccc}
		\toprule
		Model &  MRR$\uparrow$ & Hits@1$\uparrow$ & Hits@3$\uparrow$ & Hits@10$\uparrow$\\ 
        \midrule
        \ours & \textbf{38.10} & \textbf{30.13} & \textbf{42.77} & 53.59\\
        - w/o $\Ls_{tdt}$ & 37.48 & 28.43 & 42.34 & \underline{53.62}\\
        - w/o $\Ls_{sft}$ & 37.31 & 28.07 & \underline{42.64} & 53.42\\
        - w/o $\Ls_{nce}$ & 14.09 & 10.40 & 15.24 & 21.29\\
        - w/o Conditional Attention & 37.20 & 27.69 & 42.13 & 53.60\\
        - Shared Head MLP & 28.01 & 19.57 & 31.98 & 44.73\\
        - Shared Score Layer & \underline{37.54} & \underline{28.64} & 42.12 & \textbf{53.73}\\
		\bottomrule
	\end{tabular}}
        \caption{Ablation studies on the DB15K dataset. Shared Head MLP and Shared Score Layer are the methods with one shared adaption module instead of $K$ different modules. }
	\label{tab:ablation}
\end{table}

\begin{figure}[thb]
    \centering
    \includegraphics[width=\linewidth]{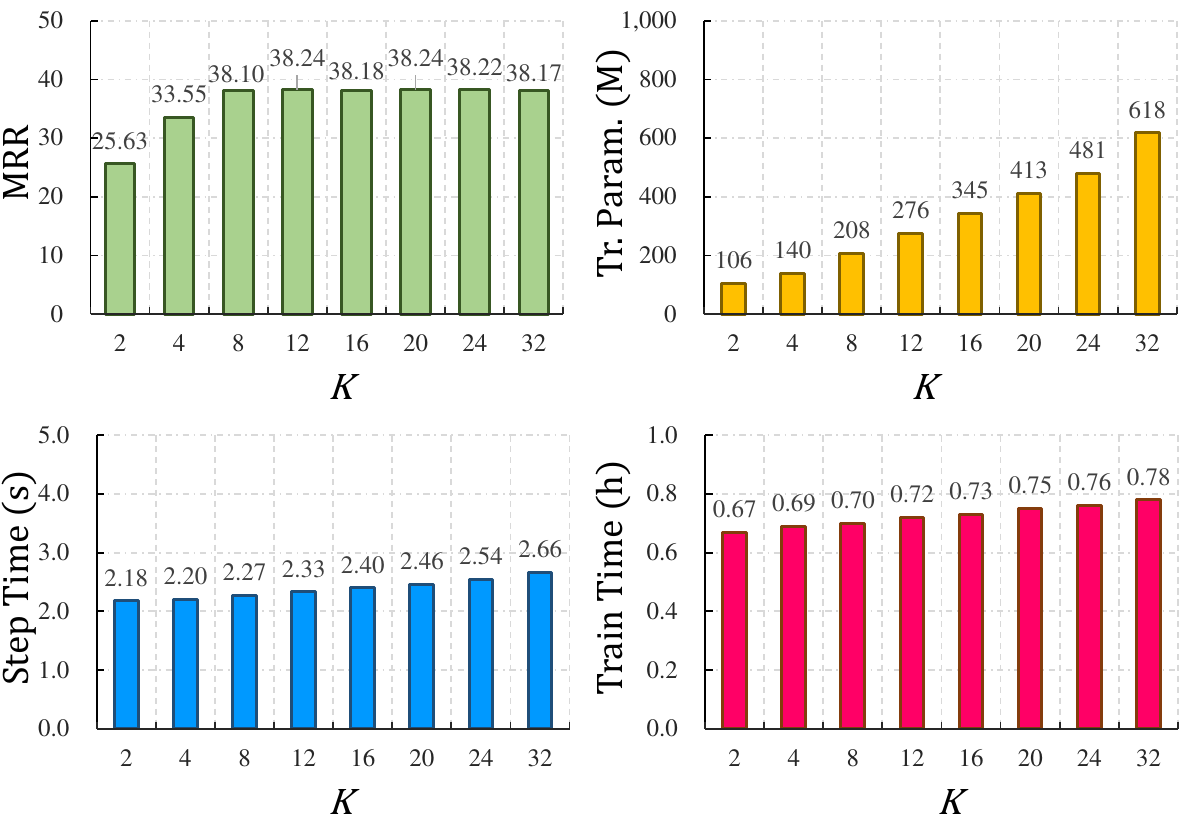}
    \caption{Performance of \ours w.r.t. the number of \ours head layers $k$. The results are obtained using $8$ A100 GPUs.}  
    \label{fig:num_kon}
\end{figure}

\begin{figure*}[thb]
    \centering
    \includegraphics[width=.95\textwidth]{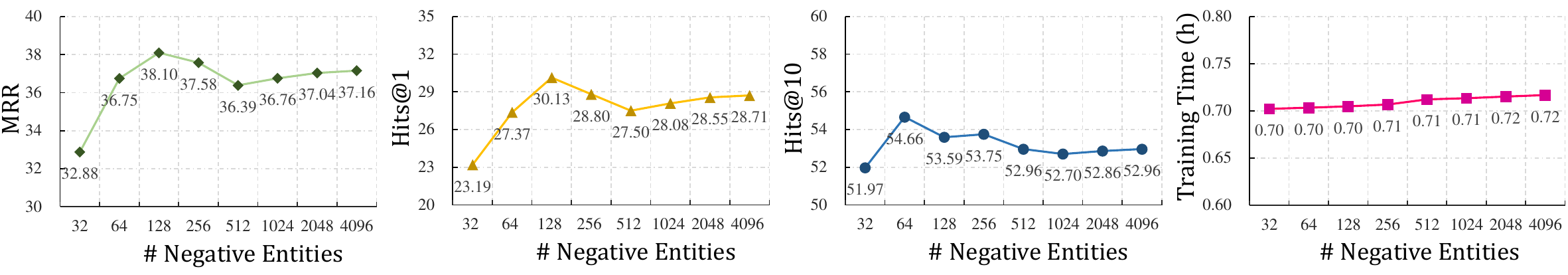}
    \caption{Performance of \ours w.r.t. the number of negative entities. The results are obtained using $8$ A100 GPUs.}  
    \label{fig:num_neg}
\end{figure*}

\subsection{Ablation Studies}
We conduct ablation studies to verify the effectiveness of each module in \ours. We remove or replace the core modules in \ours to conduct experiments. Specifically, w/o $\Ls_{tdt}$, w/o $\Ls_{sft}$, w/o $\Ls_{nce}$ refer to the methods that exclude the token distribution tuning loss, the token supervised fine-tuning loss, and the entity-level negative contrastive loss, respectively. w/o Conditional Attention is the method without the attention module. Shared Head MLP and Shared Score Layer refer to the methods where we replaced the  $K$ different LoRA layers for each step with a single shared LoRA layer for all steps. 

The results are presented in Table~\ref{tab:ablation}. w/o $\Ls_{nce}$ performs worst among all alternative methods, but it still shows some effectiveness, likely due to the presence of the token-level supervised fine-tuning loss. The substantial gap between \ours and w/o $\Ls_{nce}$ highlights the importance of entity-level contrastive learning. 
 $\Ls_{tdt}$ and $\Ls_{sft}$ are also crucial for \ours, and removing either one results in a significant performance drop in MRR, Hits@1, and Hits@10. Interestingly, we observe that the Hits@10 results are relatively insensitive to our ablation settings, showing only minimal differences across multiple methods.

Removing the conditional attention module also leads to a significant performance decline, particularly in Hits@1. We believe this module is essential for accurately identifying target entities. We also develop two shared-weight variants. The method leverages one shared score layer achieves the second-best performance, suggesting that \ours effectively shapes the $k$-step hidden states. It is similar to the original LLM that leverages one shared score layer to estimate the probability distribution of different steps. In contrast, the method with a shared head MLP suffers a significant performance drop, resulting in the second-worst performance. This empirically confirms that merely fine-tuning the score layer is insufficient for our task, even when using $k$ different LoRA adaptions.

\begin{table}[t]
	\centering
	\resizebox{\linewidth}{!}{\scriptsize
	\begin{tabular}{ll|cccccc}
		\toprule
		Operator & Weight &  MRR$\uparrow$ & Hits@1$\uparrow$ & Hits@3$\uparrow$ & Hits@10$\uparrow$\\ 
        \midrule
        \multirow{2}{*}{+} & learnable & \textbf{38.10} & \textbf{30.13} & \textbf{42.77} & \textbf{53.59}\\
         & constant & \underline{37.11} & \underline{29.08} & \underline{41.82} & \underline{52.97}\\ 
        \multirow{2}{*}{*} & learnable & 23.24 & 15.95 & 26.37 & 37.70 \\
        & constant & 23.61 & 16.36 & 26.97 & 37.93\\
		\bottomrule
	\end{tabular}}
        \caption{The results of \ours with different estimation functions for the joint entity probability. $+$ and $*$ denote the addition and multiplication operations, respectively.}
	\label{tab:joint}
\end{table}

\subsection{Analysis on the \ours Heads}
The number of \ours heads (denoted as $K$) is a crucial hyper-parameter, determining the maximum token length available for representing each entity.
While a larger $K$ can prevent the truncation of entity names and potentially enhance model performance, it also increases computational demands. To examine this trade-off, we conducted experiments with varying values of $K$.

The results are illustrated in Figure~\ref{fig:num_kon}, where we present four subgraphs depicting MRR, the number of trainable parameters, per-step time, and overall training time. Notably, performance appears to saturate when $K\geq 8$,likely because most entity names consist of fewer than 8 tokens. Nonetheless, the computational cost increases linearly with $K$. As observed, both step time and training time show slight increases with larger $K$, while the number of trainable parameters significantly rises. Therefore, we choose $K=8$ for the main experiments, as it strikes an optimal balance between performance and computational efficiency.

\subsection{Analysis on the Entity-level Contrastive Loss}

One of the key features of \ours is the entity-level contrastive loss, defined by one positive example and $|\gN|$ negative examples. It is interesting to analyze how varying $|\gN|$ impacts both the performance and computational cost of \ours. As illustrated in Figure~\ref{fig:num_neg}, increasing the number of negative examples does not necessarily lead to improved performance. Our observations reveal that the performance curves across all three metrics exhibit a similar trend: initially, there is a rapid increase, followed by a steady decline, and ultimately, a plateau phase.

In contrast to the effects of varying the number of head layers $K$, we find that the computational cost remains relatively stable even as $|\gN|$ increases. This stability arises because the negative examples are extracted from the 
$K$ head output probability distributions of \ours. Thus, adding more negative examples does not involve additional neural layers. Based on these findings, we have selected $|\gN|=128$ as the optimal setting for our main experiments.

\subsection{Analysis on the Joint Probability Function}
The entity-level contrastive loss requires the entity probability as input, which is derived from its named tokens. As such, selecting an appropriate method to combine these token probabilities into a joint probability for the entity is crucial. We investigate four different methods for estimating this joint probability. The operations of addition and multiplication are denoted by $+$ and $*$, respectively. The term \emph{learnable} refers to the version where we utilize a weight vector to aggregate the probabilities of the tokens at different positions, which is shared across entities. Conversely, \emph{constant} indicates an unweighted aggregation.

Table~\ref{tab:joint} illustrates the results on the DB15K dataset. Notably, the learnable $+$ method achieves the highest performance across all metrics. When we remove the learnable weights, there is a slight degradation in performance. In contrast, although the $*$ operator appears to be intuitively more effective, it significantly underperforms compared to $+$ across all metrics. We suspect that the lackluster performance of the multiplication method may be attributed to the issue of vanishing gradients. Specifically, multiplying multiple ($8$ in our experiments) probabilities may result in exceedingly small scalar values, although conceptually valid as a joint probability, might hinder the learning process.

\section{Conclusion and Limitation}
\label{sec:concl}
In this paper, we propose \ours to stack the knowledge on the head layer of LLM. We introduce an entity-level contrastive loss that significantly reduces computational costs and develope HTT to align the output distributions with those of the original LLM head. Extensive experiments demonstrate the superior performance of \ours compared to state-of-the-art baselines. \ours still has limitations. First, its flexibility is constrained, as it does not support arbitrarily large values of $K$. To address this, we plan to explore a sliding window mechanism for processing $K$-step prediction in future; Second, \ours currently lacks support for multi-modal input. We also intend to incorporate large vision-language models into \ours as part of our future work.

\section*{Acknowledgements}
We would like to thank all anonymous reviewers for their insightful and invaluable comments. This work is founded by National Natural Science Foundation of China (NSFCU23B2055/NSFC62306276/NSFCU19B2027), Zhejiang Provincial Natural Science Foundation of China (No. LQ23F020017), Yongjiang Talent Introduction Programme (2022A-238-G), and Fundamental Research Funds for the Central Universities (226-2023-00138). This work was supported by AntGroup.

\bibliography{references}

\end{document}